\documentclass[11pt,letterpaper]{article}
\usepackage[margin=0.75in]{geometry}
\usepackage[T1]{fontenc}
\usepackage{lmodern}
\usepackage[hyphens]{url}
\usepackage{graphicx}
\usepackage{float}
\usepackage{amsmath}
\usepackage{amssymb}
\usepackage{natbib}
\usepackage{caption}
\usepackage{textcomp}
\usepackage{booktabs}
\usepackage{enumitem}
\usepackage{xcolor}
\usepackage{hyperref}
\setlist[itemize]{leftmargin=*, topsep=2pt, itemsep=2pt}
\title{FG-ConvNO: A Geometry-Aware Neural Operator for Propeller CFD Prediction}
\author{%
Yichen Di\textsuperscript{1},
Valentin Duruisseaux\textsuperscript{2},
Di Zhou\textsuperscript{3},
Xinyi Li\textsuperscript{2},\\
Daniel Leibovici\textsuperscript{4},
Jean Kossaifi\textsuperscript{4},
Anima Anandkumar\textsuperscript{2}\\[0.7em]
\small \textsuperscript{1}Tsinghua University \quad
\textsuperscript{2}California Institute of Technology\\
\small \textsuperscript{3}University of Tennessee \quad
\textsuperscript{4}NVIDIA\\[0.4em]
\small \texttt{diyc041018@gmail.com},
\texttt{vduruiss@caltech.edu},
\texttt{xinyili@caltech.edu}
}
\date{}

\begin{document}
\maketitle

\begin{abstract}
Computational Fluid Dynamics (CFD) is essential in contemporary engineering design, yet it remains expensive to compute and can take thousands of CPU/GPU hours or more for a single high-fidelity simulation. Recent developments in machine learning for scientific modeling have led to powerful surrogate frameworks that can approximate flow fields and physical quantities by using a fraction of the computational cost. However, most of these methods require large-scale homogeneous datasets or simple geometries, exhibiting severe generalization gaps when applied to complex and data-scarce configurations. To address this, we propose the Factorized Geometry Convolutional Neural Operator (FG-ConvNO) for robust geometry-aware propeller pressure prediction under limited training data. We introduce geometry-aware encoding, efficient interpolation-based decoding, and auxiliary global quantity supervision. In addition, we incorporate DISCO blocks that use discrete continuous convolutions to make the model discretization agnostic. In our experiments, FG-ConvNO achieves the highest accuracy for both surface pressure and global thrust prediction on a procedurally generated propeller CFD dataset. Geometry-aware encoding and architectural simplification improve stability, while random vertex subsampling enhances robustness. Overall, adding geometric priors to grid-based convolutional operators reduces overfitting on complex propeller geometries and improves generalization in low-data regimes.

\end{abstract}

\hfill 

\section{Introduction}

\textbf{Computational Fluid Dynamics (CFD)} is the cornerstone of modern aerospace and engineering, enabling accurate prediction of aerodynamic and hydrodynamic behavior. However, traditional CFD solvers are computationally prohibitive: resolving turbulent 3D flows often requires thousands of GPU hours or more for a single geometry \citep{10.1145/3197517.3201325}. This cost severely limits rapid design iterations, parameter study, and uncertainty estimation for industrial applications. Consequently, there has been an increasing interest in machine-learning-based surrogates that efficiently emulate CFD solvers while preserving physical fidelity \citep{wang2024recentadvancesmachinelearning}, using architectures ranging from convolutional and graph neural networks to neural operators.

 Among them, neural operators are a class of methods that can learn mappings from boundary conditions and geometric representations to continuous solution fields~\citep{kovachki2023neural,azizzadenesheli2024neural}. Architectures such as the Fourier Neural Operator (FNO) \citep{li2021fourierneuraloperatorparametric}, the Graph Neural Operator (GNO) \citep{li2020neuraloperatorgraphkernel}, and the Geometry-Informed Neural Operator (GINO) \citep{li2023geometryinformedneuraloperatorlargescale} enable resolution-independent inference, learning PDE solution operators rather than fixed discretized mappings~\citep{Berner2025Principles}. Recent advances also include DISCO \citep{liuschiaffini2024neuraloperatorslocalizedintegral}, which replaces traditional convolution kernels with continuous kernels to help stabilize operator behavior under resolution changes. These models demonstrate exceptional performance on various benchmarks, including physical systems involving elasticity, plasticity, and turbulent flow systems such as airfoils and pipes. They also excel on aerodynamic datasets such as ShapeNet-Car \citep{chang2015shapenetinformationrich3dmodel}, which feature smooth, structured geometries.

Even though these methods can deal with unstructured point clouds, they typically require access to extensive datasets to learn from. Their generalization capabilities to unseen shapes and geometries can deteriorate significantly when training data is scarce, especially for intricate designs like propellers. There, the base geometries are highly nonlinear with complex curvature, skew and pitch changes leading to very nonuniform pressure distributions. In these contexts, models trained on a few hundred objects often suffer from two competing failure modes: underfitting and overfitting. 

\vspace{2mm}

\textbf{Underfitting.}  
In the limited data setting, diverse geometries often cause data-driven models to collapse toward nearly uniform surface predictions. Sharp variations at tips or hub regions become overly smoothed, indicating that the model fails to learn the nonlinear geometry–flow relationship and instead defaults to global averages.

\textbf{Overfitting.}
With limited data, high-capacity models fit the training set quickly but fail to generalize. The test loss plateaus early, showing that the network memorizes sample-specific noise instead of learning stable flow patterns. This appears as noisy surface predictions where local flow features are memorized rather than generalized.

\textbf{Proposed Approach.}  
We propose a new architecture, the Factorized Geometry Convolutional Operator (FG-ConvNO), based on the Factorized Implicit Global Convolution (FIGConv) but reformulated for robustness and geometry-awareness under limited CFD data.

\begin{itemize}
    \item \textbf{Simplified architecture:} FG-ConvNO  retains the hierarchical design from FIGConv but reduces its depth and increases the number of blocks per level, avoiding overfitting while maintaining a sufficiently large receptive field. In addition, the grid-to-vertex decoding is reformulated via trilinear interpolation, yielding smoother feature reconstruction and lower memory consumption.
    
    \item \textbf{Geometry-aware encoding:}  
    We combine signed distance functions (SDFs), surface normals, and boundary condition embeddings in the vertex–grid exchange layers. This coupling enables FG-ConvNO to utilize geometric context and flow boundary information in more ways.

    \item \textbf{Integration of DISCO blocks:} In FG-ConvNO, we replace standard CNN blocks with DISCO blocks that perform discrete continuous convolution and  maintain a consistent receptive field across different grid resolutions, enabling discretization invariance.

    \item \textbf{Global prediction objective:}  
    The coarsest latent grid is pooled to regress global aerodynamic or hydrodynamic quantities, e.g., thrust that brings integral physical supervision from a local pressure fields to global performance metrics.
    
\end{itemize}

We also introduce a new (small) dataset of RANS simulations for diverse propeller geometries. 

\textbf{Summary of Results.} Our numerical experiments on the newly introduced propeller CFD dataset demonstrate the effectiveness of the proposed modifications:

\begin{itemize}
    \item \textbf{Superior surface accuracy and robustness:}  
    FG-ConvNO achieves the lowest surface pressure relative error (23.68\% with surface-only training and 24.07\% with joint surface and global supervision), outperforming GINO (29.16\%), Transolver++ (27.96\%), and GraphSAGE (42.47\%).

    \vspace{2mm}
    
    \item \textbf{Physically consistent global prediction:}  
    Explicit supervision of global quantities (e.g. total thrust and torque) yields the best thrust estimates, with a 7-9$\times$ reduction in global loss compared to GINO and Transolver++ models.

     \vspace{2mm}
     
    \item \textbf{Decoder and SDF ablation:} 
    Decoding using interpolation leads to more robust results, while the incorporation of signed distance functions (SDFs) improves geometric consistency. The combined setup achieves the best compromise between accuracy and physical consistency.

\end{itemize}

\section{Related Works}

\textbf{Graph and Grid-Based Neural Operators.}
Early CFD surrogates in geometric domains naturally adopted graph neural networks (GNNs)  such as GraphSAGE \citep{hamilton2018inductiverepresentationlearninglarge} and MeshGraphNet \citep{pfaff2021learningmeshbasedsimulationgraph}. 
However, they rely on hard-coded adjacency, and denser sampling may cause message passing to degenerate into a single-point operator. The Graph Neural Operator (GNO) \citep{li2020neuraloperatorgraphkernel} mitigates this by using radius-based aggregation to approximate local kernel integration in continuous space. For regular meshes, the Fourier Neural Operator (FNO) \citep{li2021fourierneuraloperatorparametric} performs global mixing via FFTs and achieves strong resolution-invariant performance.

The Geometry-Informed Neural Operator (GINO) \citep{li2023geometryinformedneuraloperatorlargescale} 
combines the strengths of GNO and FNO by mapping irregular meshes to a latent grid for spectral convolution and projecting the results back. Building on this idea, the Factorized Implicit Global Convolution (FIGConv) 
\citep{choy2025factorizedimplicitglobalconvolution} factorizes the latent domain into anisotropic grids, reducing memory from cubic to quadratic order and replacing spectral layers with efficient 3D convolutions. This latent-space based strategy with domain factorization is particularly promising as the building block for scalable neural architectures for complex CFD applications.

\textbf{Transformer-based architectures.} 
Transolver~\citep{wu2024transolverfasttransformersolver} introduces physics-domain tokenization and self-attention over geometric tokens, while Transolver++~\citep{luo2025transolveraccurateneuralsolver} improves efficiency through memory optimization and parallelism, supporting training on millions-scale datasets.
AB-UPT~\citep{alkin2025abuptscalingneuralcfd} adopts an anchor-based design, compressing spatial information through anchor points and using cross-attention between surface and volume samples for efficient coupling.

\textbf{Physics-Informed Neural Operators and Learning-Augmented Integrators.}
Beyond purely data-driven surrogates, multiple works integrate physics or numerical priors into operator learning. Physics-Informed Neural Operators (PINOs) \citep{li2023physicsinformedneuraloperatorlearning,lin2025mGNO, ganeshram2025fcpinohighprecisionphysicsinformed} extend the idea of Physics-Informed Neural Networks (PINNs) \citep{RAISSI2019686} to the operator learning paradigm by fusing coarse-resolution data with high-resolution PDE constraints, enabling super-resolution inference and better parametric stability. In parallel, the Neural Operator Element Method (NOEM) \citep{ouyang2025neuraloperatorelementmethodefficient} embeds neural operators inside the FEM pipeline to reduce meshing cost in complex subdomains. 

\section{Methods}

\subsection{Dataset: Propeller CFD}

The dataset consists of procedurally generated 3D propellers simulated using the open-source finite-volume solver OpenFOAM\textregistered. Each case is parameterized by the blades' pitch, skew and chord length at selected radial positions, from which a complete mesh is generated and evaluated under physical conditions defined by the inflow velocity and angular speed $\omega$. To model rotating propellers, the Multiple Reference Frame (MRF) technique is employed, providing a computationally efficient way to account for the rotating frame of reference. High-performance parallel computing is used to perform Reynolds-Averaged Navier–Stokes (RANS) simulations for these propeller configurations, producing steady-state flow quantities including pressure, wall shear stress, and integrated forces (thrust and torque).

Each sample is represented as $\{v, c, u, g\}$, where $v \in \mathbb{R}^{n\times3}$ are the spatial coordinates of the point cloud, $c \in \mathbb{R}^{d_c}$ encodes global boundary conditions, $u \in \mathbb{R}^{n\times d_u}$ denotes local quantities such as pressure and wall shear stress, and $g \in \mathbb{R}^{d_g}$ global quantities such as total thrust or torque. Surface prediction serves as the primary objective for capturing detailed flow behavior, while accurate global prediction is equally critical for downstream applications such as performance evaluation and geometry optimization.  
The dataset includes approximately 1000 propeller cases, each containing $\sim$70k surface points and million-scale volume points.

\begin{figure}[H]
    \centering
    \includegraphics[width=0.5\linewidth, height=0.22\textheight, keepaspectratio]{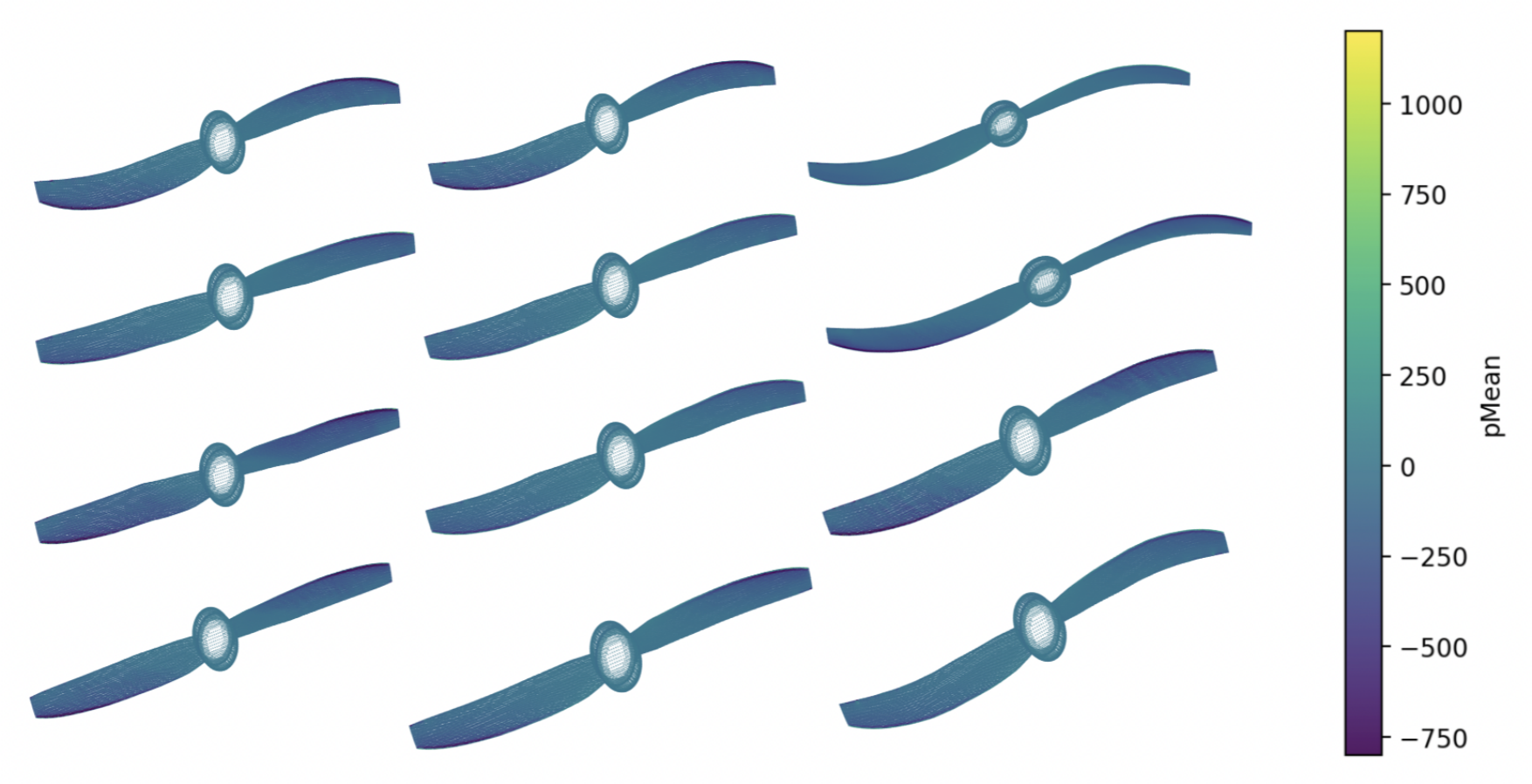} \vspace{-1mm}
    \caption{
    \textbf{Visualization of procedurally generated propeller geometries.}
    Each sample is colored by time-averaged pressure obtained from simulations, illustrating flow variations across different blade designs.}
    \label{fig:propeller_samples}
\end{figure}

\subsection{Factorized Implicit Global Convolutions}

The Factorized Implicit Global Convolution (FIGConv)~\citep{choy2025factorizedimplicitglobalconvolution} is a CNN-based neural network designed to learn continuous 3D mappings. FIGConv creates a set of shared factorized latent grids $\{F_m\}_{m=1}^{M}$ for all training cases. Each grid is defined on two high-resolution axes and one low-resolution axis, effectively decomposing a dense 3D volume into several anisotropic 2.5D slices. For instance, $F_1$ may have dimensions $(r, W_{\max}, D_{\max})$, $F_2$ as $(H_{\max}, r, D_{\max})$, and $F_3$ as $(H_{\max}, W_{\max}, r)$, where $r \ll H_{\max},W_{\max},D_{\max}$.
This factorization reduces voxels from cubic to roughly quadratic scale while preserving full 3D coverage. The vertex–grid encoder maps surface points onto these grids, each capturing structure along two dense axes, and their combined features reconstruct the full field.

\begin{figure}[H] \vspace{-2mm}
    \centering
\includegraphics[width=\linewidth, height=0.55\textheight, keepaspectratio]{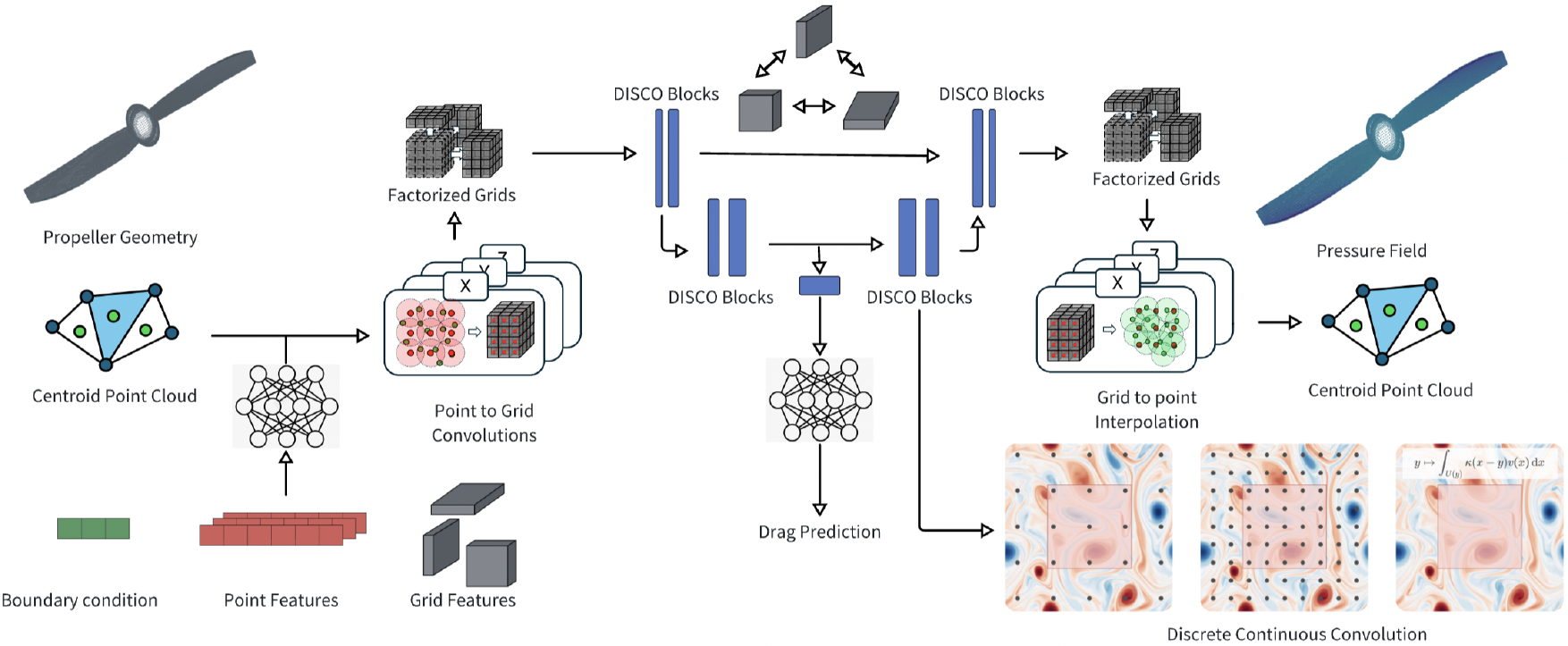} \vspace{-1mm}
    \caption{
    \textbf{The Factorized Geometry Convolutional Neural Operator (FG-ConvNO) for propeller pressure prediction.}
    Each propeller geometry is first represented as a centroid-based surface point cloud, where each centroid corresponds to a local patch on the propeller surface. FG-ConvNO then projects vertex features onto multiple latent grids, performs hierarchical convolutional aggregation across the grids, and decodes the resulting latent fields back to the vertex domain for pressure prediction. Compared with FIGConv, FG-ConvNO incorporates geometry-aware conditioning (using boundary conditions, SDFs, and normals), interpolation-based feature decoding, an explicit global prediction head to infer quantities such as thrust and torque, and hierarchical DISCO blocks for discretization convergence, which preserves consistent local support across different grid resolutions.}
    \label{fig:fgnet_arch} \vspace{-1mm}
\end{figure}

\paragraph{Positional Encoding.}
FIGConv encodes all spatial coordinates using sinusoidal positional embeddings. These encodings provide translational and rotational consistency by representing relative offsets as frequency-domain phase shifts rather than absolute coordinates. Consequently, neighboring vertices and grid voxels exhibit smoothly varying embeddings, enabling coherent interpolation between the irregular vertex domain and the structured latent grids.

\paragraph{Vertex--Grid Encoder and Decoder.}
The encoder and decoder adopt a point convolution formulation for vertex--grid feature exchange.
For encoding, each voxel center $p_{m,ijk}$ in grid $F_m$ aggregates the features of its neighbors:
\begin{align*}
f_{m,ijk} 
&= \sum_{v \in \mathcal{N}(p_{m,ijk})}
    \psi_m\big(\phi(v), \phi(p_{m,ijk}), \phi(v - p_{m,ijk})\big) \\
&\approx \int_{\mathcal{N}(p_{m,ijk})} 
    \kappa_m(v, p_{m,ijk})\, \rho(v)\, dv.
\end{align*}
Here, $$\mathcal{N}(p_{m,ijk})= \{\, v \mid \| \Sigma_m^{-1/2}(v - p_{m,ijk}) \| < 1 \,\}$$ defines an ellipsoidal neighborhood centered on $p_{m,ijk}$ and determined by a covariance matrix $\Sigma_m$. This is similar to the radius-based aggregation used in GNO, but with an ellipsoidal (rather than spherical) region that aligns with the directional structure of each factorized grid. The function $\psi_m(\cdot)$ implements the kernel $\kappa_m$ through an MLP, and $\rho(v)$ represents the density function of the vertex. 
This continuous formulation allows the encoder to integrate geometric information from irregular vertex distributions while preserving local spatial coherence. The decoder mirrors this process in the reverse direction, reconstructing vertex embeddings from nearby voxel features,
\[
\hat{f}(v) =  \! \!\sum_{p_{m,ijk}\in \mathcal{N}(v)}  \!
\psi'_m\big(\hat{f}_{m,ijk}, \phi(p_{m,ijk}), \phi(v - p_{m,ijk})\big).
\]
This symmetric design ensures consistent information flow between the physical domain and the structured latent grid.

\paragraph{Latent Grid Convolution.}
All latent grids $\{F_m\}$ are jointly processed by a U-Net backbone composed of convolution, downsampling, and upsampling stages.
Each 3D convolution applies a local kernel $W_m$ to grid $F_m$, followed by inter-grid feature exchange that allows communication across different orientations:
\begin{align*}
\tilde{F}_m^{(l)} \ & =  
\mathrm{Conv3D}(F_m^{(l)}; W_m) \\
F_m^{(l+1)} &= \tilde{F}_m^{(l)} + \sum_{m'\neq m} 
\mathrm{Interp}\big(\tilde{F}_{m'}^{(l)}, \text{coords}(F_m)\big).
\end{align*}
The fusion step enables the propagation of information across all factorized grids. 

\subsection{Factorized Geometry Convolutional Neural Operator}

We extend FIGConv to the Factorized Geometry Convolutional Neural Operator (FG-ConvNO), designed to improve generalization under data scarcity and complex geometry. Its baseline configuration follows FIGConv, where only surface vertices $v_\text{s}$ are used as input, and training is supervised by a $\ell_2$ relative loss in surface pressure predictions.

\paragraph{Architectural Simplification.}
To alleviate overfitting and improve stability, FG-ConvNO reformulates FIGConv by simplifying both its hierarchical structure and the vertex–grid interaction mechanism.
Specifically, we retain the multi-scale U-Net–like design from FIGConv but reduce its depth ($n_{\text{levels}}$) while increasing the number of convolution blocks per level ($n_{\text{down}}$, $n_{\text{up}}$). This heuristic is used to maintain the receptive field in the new configuration close to that of its original one at a lower information loss due to redundant downsampling. Second, we replace the grid-to-vertex point convolution in the decoder with an interpolation-based feature exchange implemented using trilinear sampling. For each output vertex $v$, the features are interpolated from all factorized grids through differentiable sampling:
\[
\hat{f}(v) = \sum_{m=1}^{M} 
w_m \cdot 
\mathrm{GridSample}\!\left(F_m, \mathrm{proj}(v)\right),
\]
where $\mathrm{GridSample}(\cdot)$ denotes trilinear interpolation, $\mathrm{proj}(v)$ maps $v$ to the normalized coordinates of grid $F_m$, and $w_m$ are learned per-grid weights. 
This interpolation avoids repeated kernel evaluations, significantly reducing memory consumption, and ensures spatial smoothness of the reconstructed pressure field since the interpolated features continuously vary with vertex position.

\paragraph{Geometry-Aware Encoding.}
While FIGConv only uses positional encoding for grid feature information, FG-ConvNO pursues geometric and physical awareness by exploiting multiple conditioning signals such as boundary conditions, signed distance functions (SDFs) and surface normals. We integrate them through a learned feature fusion module $\mathrm{Fuse}(\cdot)$, which can be weighted summation or attention-based mixing. For each surface vertex $v$ and grid point $p_{m,ijk}$, we first compute their respective embeddings and then aggregate them via the fusion module:
\begin{align*}
\tilde{\phi}(v) 
&= \mathrm{Fuse}\big(\phi(v), \gamma(c), \gamma(n_v)\big), \\
\tilde{\phi}(p_{m,ijk}) 
&= \mathrm{Fuse}\big(\phi(p_{m,ijk}), \gamma(\mathrm{SDF}(p_{m,ijk}))\big).
\end{align*}
This enhanced representation enables FG-ConvNO to capture geometry and boundary-aware correlations more effectively than FIGConv, improving feature consistency and generalization across diverse propeller geometries.

\begin{figure}[H]
    \centering
    \includegraphics[width=0.6\linewidth, height=0.28\textheight, keepaspectratio]{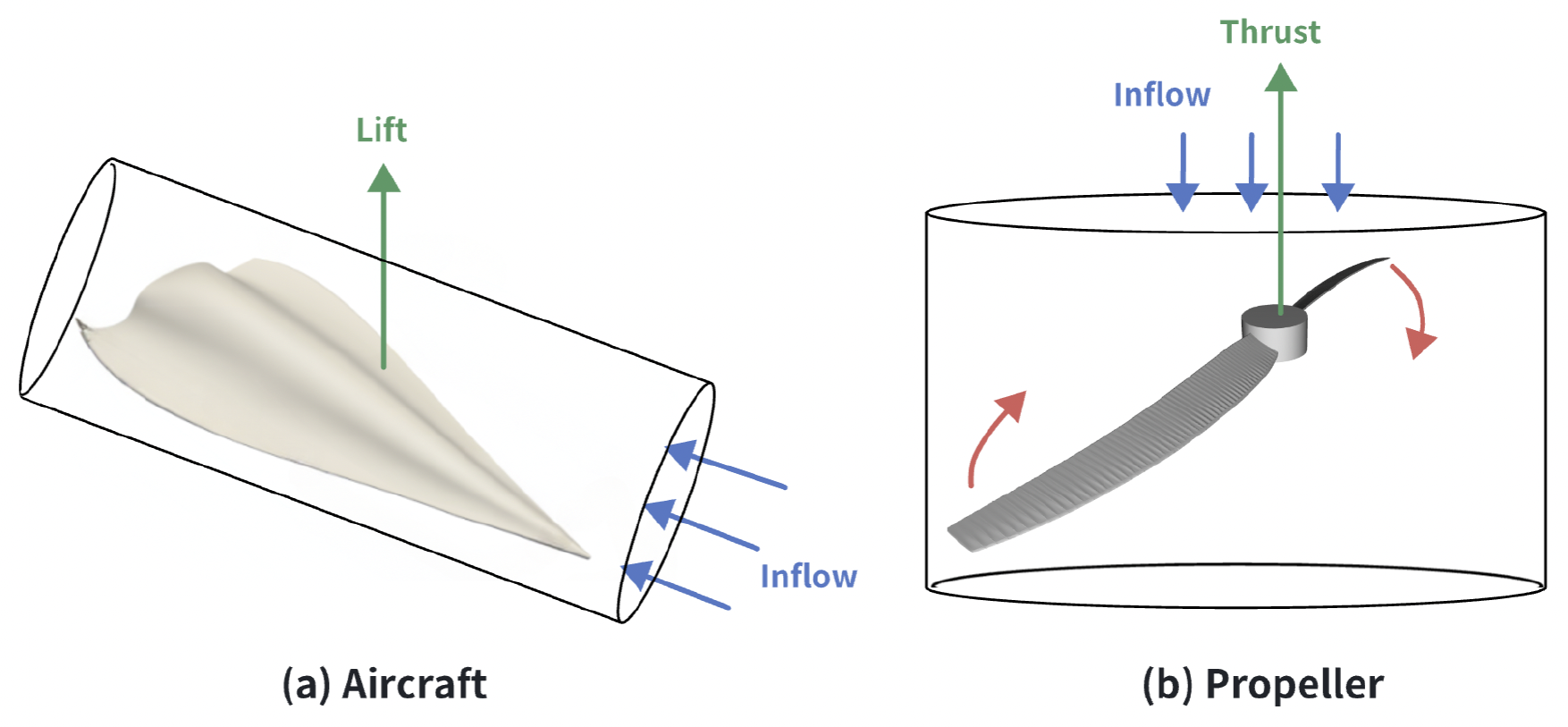}
    \caption{
    \textbf{Comparison of aerodynamic and propulsive forces.}
    (a) Aircraft wings generate lift perpendicular to the inflow. (b) Propellers produce thrust opposite to the inflow. This illustrates how global quantities such as thrust and drag integrate local pressure distributions.}

    \label{fig:global_prediction}
\end{figure}

\paragraph{DISCO Blocks}
A standard convolution assumes that the kernel acts on grid indices rather than geometric coordinates.  
For a regular grid \(D_h=\{z_i\}\) with spacing \(h\), a discrete convolution takes the form
\[
\operatorname{Conv}_{K}[v](z_i)
   =\sum_{t=1}^{S} K_t\, v(z_i + z_t),
\]
where \(z_t\) denotes a fixed stencil offset such as \((\pm h,0)\) or \((\pm h,\pm h)\).  
Both the neighbor set \(\{z_i+z_t\}\) and the kernel coefficients \(K_t\) are tied to this discretization.  
As the grid is refined, these offsets shrink, the receptive field collapses, and the convolution converges to a pointwise operator, so CNN kernels depend inherently on the chosen resolution.

DISCO replaces this with a continuous formulation
\[
(\kappa * g)(v)=\int_{B_r(v)} \kappa(u-v)\, g(u)\, du,
\]
where neighbors come from a radius-based set \(B_r(v)\) independent of discretization.  
The kernel is represented as
\[
\kappa(z)=\sum_{\ell=1}^{L}\theta_\ell\, \phi_\ell(z),
\]
using continuous basis functions \(\phi_\ell\) and trainable coefficients \(\theta_\ell\).  
The integral is approximated by quadrature,
\[
(\kappa * g)(v_i)\approx \sum_{u_j\in B_r(v_i)}\kappa(u_j-v_i)\, g(u_j)\, q_j,
\]
so the operator evaluates continuous offsets rather than discrete grid steps.  
This yields a stable receptive field and makes the kernel reusable across resolutions.

In FG-ConvNO, every CNN block used in FIGConv is replaced with a DISCO block while keeping the same intra-grid communication. At each downsampling level, the latent grid resolution is halved, while the DISCO convolution radius is doubled, which ensures that each layer represents the same continuous operator despite operating on grids of different resolutions. Down blocks therefore consist of a DISCO convolution followed by average pooling for spatial halving, while up blocks apply bilinear upsampling before the DISCO convolution to restore spatial resolution. With this construction, the network maintains consistent local support and discretization invariance.

\subsection{Global Prediction Objective}
The local surface pressure is a measure of the fine grained flow behavior, but as far as engineering system design is concerned, the integrated global response variables (total thrust and total torque) matter the most, as they drive the performance of a propulsion device. These global forces are physically analogous to lift and drag on aircraft wings or automotive surfaces (Fig.~\ref{fig:global_prediction}), where distributed pressure fields integrate into macroscopic aerodynamic effects.

In FG-ConvNO, we predict such global quantities (e.g., thrust, torque, efficiency) directly, extending FIGConv with explicit global physical supervision. The lowest-resolution latent grid within the U-Net hierarchy serves as a compact representation of the entire flow field, from which the network infers global variables $g$ summarizing the overall aerodynamic or hydrodynamic performance of the propeller.
Formally, these quantities correspond to surface integrals of local flow variables,
\[
g = \int_{S} \big(p(v)\, n_v + \tau(v)\big)\, dS,
\]
where $p(v)$ is the surface pressure, $n_v$ the normal vectors and $\tau(v)$ the shear-stress component. 
In practice, FG-ConvNO does not perform this integration explicitly; instead, it learns to approximate this mapping directly using the feature maps of the coarsest latent grids $\{F_m^{(L)}\}$ :
\[
z_\text{global} = \mathrm{Pool}\big(\{F_m^{(L)}\}\big), \quad
\hat{g} = \mathrm{MLP}(z_\text{global}).
\]
The training loss combines global MSE $\mathcal{L}_\text{global} = \|\, \hat{g} - g \,\|_2^2$ and local surface supervision, i.e., 
\[
\mathcal{L} =
\mathcal{L}_\text{rel}(y_\text{s}, \hat{y}_\text{s})
+ \lambda_\text{g}\, \|\, \hat{g} - g \,\|_2^2,
\]
where both $g$ and $y_s$ are normalized with respect to their own statistical ranges during training to ensure scale consistency. This joint formulation links local pressure prediction with physically significant global quantities, improving overall physical consistency. This global supervision mechanism is model-agnostic and can be readily applied to other surrogate settings where dense fields must remain consistent with integral physical quantities.

\section{Experiments}

\subsection{Experimental Setup}

All experiments are conducted on the proposed propeller CFD dataset described in Section~3.1. 
Unless otherwise stated, the training set contains 800 samples and the test set 200 samples. 
Each sample includes surface vertices with the corresponding normals, local flow quantities, and global responses such as thrust and torque. 
Optimization uses Adam with a learning rate of $1\times10^{-3}$, a StepLR scheduler (step size 25, decay factor $\gamma=0.2$), and batch size 1 for 200 epochs. We compare our method against representative neural operator architectures, including graph-based GraphSAGE~\citep{hamilton2018inductiverepresentationlearninglarge}, and two recent neural operator frameworks, GINO~\citep{li2023geometryinformedneuraloperatorlargescale,kossaifi2024neural} and Transolver++~\citep{luo2025transolveraccurateneuralsolver}.
All models receive identical geometric and boundary condition inputs and have comparable parameter counts for fair comparison.

Both FG-ConvNO variants achieve lower surface error than the graph-based baseline (GraphSAGE) and the recent operator models (GINO, Transolver++), while explicit global supervision substantially improves the accuracy of global quantity prediction.
All models are implemented in PyTorch and trained on NVIDIA A100 GPUs (32GB memory).

\subsection{Experimental results}

We compare FG-ConvNO with three representative baselines: GraphSAGE, GINO, and Transolver++.
All models are trained using only the relative surface loss $\mathcal{L}_\text{rel}$ unless otherwise noted, and their global quantities are computed by integrating the predicted surface pressure.
\[
\hat{g} = \int_{S} \hat{p}(v)\, n_v \, dS.
\]
For FG-ConvNO, we evaluate two variants:
(1) \textit{surface-only} training using $\mathcal{L}_\text{rel}$, and
(2) \textit{joint} training with both $\mathcal{L}_\text{rel}$ and $\mathcal{L}_\text{global}$.
All variants include geometry-aware inputs (normals, boundary conditions, and grid SDFs).

\begin{table}[H]
\caption{
\textbf{Performance comparison with existing architectures on the propeller dataset.}
Both FG-ConvNO variants achieve lower surface error than the graph-based baseline (GraphSAGE) and the recent operator models (GINO and Transolver++), while explicit global supervision substantially improves the accuracy of global quantity prediction.
}
\label{tab:baseline_comparison}
\vspace{-2mm}
\centering
\renewcommand{\arraystretch}{1.1}
\begin{tabular}{lcc}
\toprule
\textbf{Model} & \textbf{Surface $\downarrow$} & \textbf{Global $\downarrow$} \\
\midrule
GraphSAGE & 42.47\% & $8.32\times 10^{-3}$ \\
Transolver++ & 27.96\% & $2.41\times 10^{-3}$ \\
GINO & 29.16\% & $3.15\times 10^{-3}$ \\
\midrule
\textbf{FG-ConvNO (surface only)} & \textbf{23.68\%} & $1.76\times 10^{-3}$ \\
\textbf{FG-ConvNO (surface + global)} & 24.07\% & $\boldsymbol{3.43\times 10^{-4}}$ \\
\bottomrule
\end{tabular}
\renewcommand{\arraystretch}{1.0}
\vspace{-1.5mm}
\end{table}
 Table~\ref{tab:baseline_comparison} shows that FG-ConvNO consistently outperforms other recent architectures (GraphSAGE, GINO and Transolver++) in surface prediction, and achieves the lowest overall global loss when trained with the global objective.
This shows that incorporating the global loss $\mathcal{L}_\text{global}$ effectively couples local pressure estimation with integral physical consistency.

\subsection{Ablation Studies}

To gain better insight into the effect of each design component, we perform comprehensive ablation studies to investigate the roles played by data scale, random vertex subsampling, and network architecture choice. This provides further insights into the robustness and generalization properties of the proposed approach in different data regimes.

\begin{figure}[H]
    \centering
    \includegraphics[width=0.8\linewidth, height=0.5\textheight, keepaspectratio]{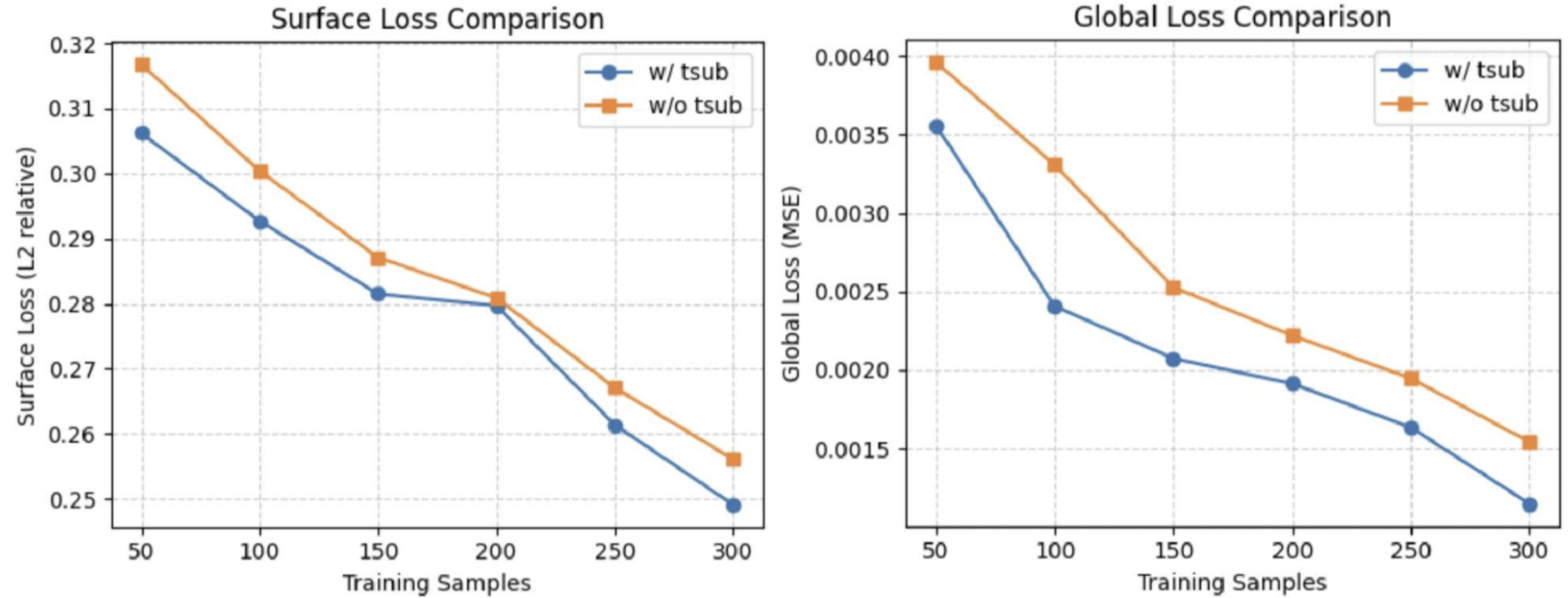} \vspace{-2mm}
    \caption{
    \textbf{Ablation on dataset size and random vertex subsampling for training (tsub)}. Increasing training samples reduces both surface and global losses, confirming the importance of data scale. Random subsampling consistently improves global prediction and provides stronger gains on surface accuracy when training data are limited.} \label{fig:data_ablation} \vspace{-3mm}
\end{figure}

\paragraph{Data Size and Random Sampling.}
We train FG-ConvNO (surface + global) using different dataset sizes (50–300 samples), with and without random vertex subsampling. 
As shown in Fig.~\ref{fig:data_ablation}, both local and global losses decrease steadily as the training set grows, demonstrating the importance of data scale for learning complex propeller geometries. Random vertex subsampling further improves generalization by exposing the model to varied point distributions, reducing dependence on mesh-specific patterns. These results highlight that increasing geometric diversity—either through more samples or sampling augmentation—is crucial for stable surrogate modeling in data-scarce settings.

\begin{figure}[H] \vspace{-2mm}
    \centering
    \includegraphics[width=\linewidth, height=0.3\textheight]{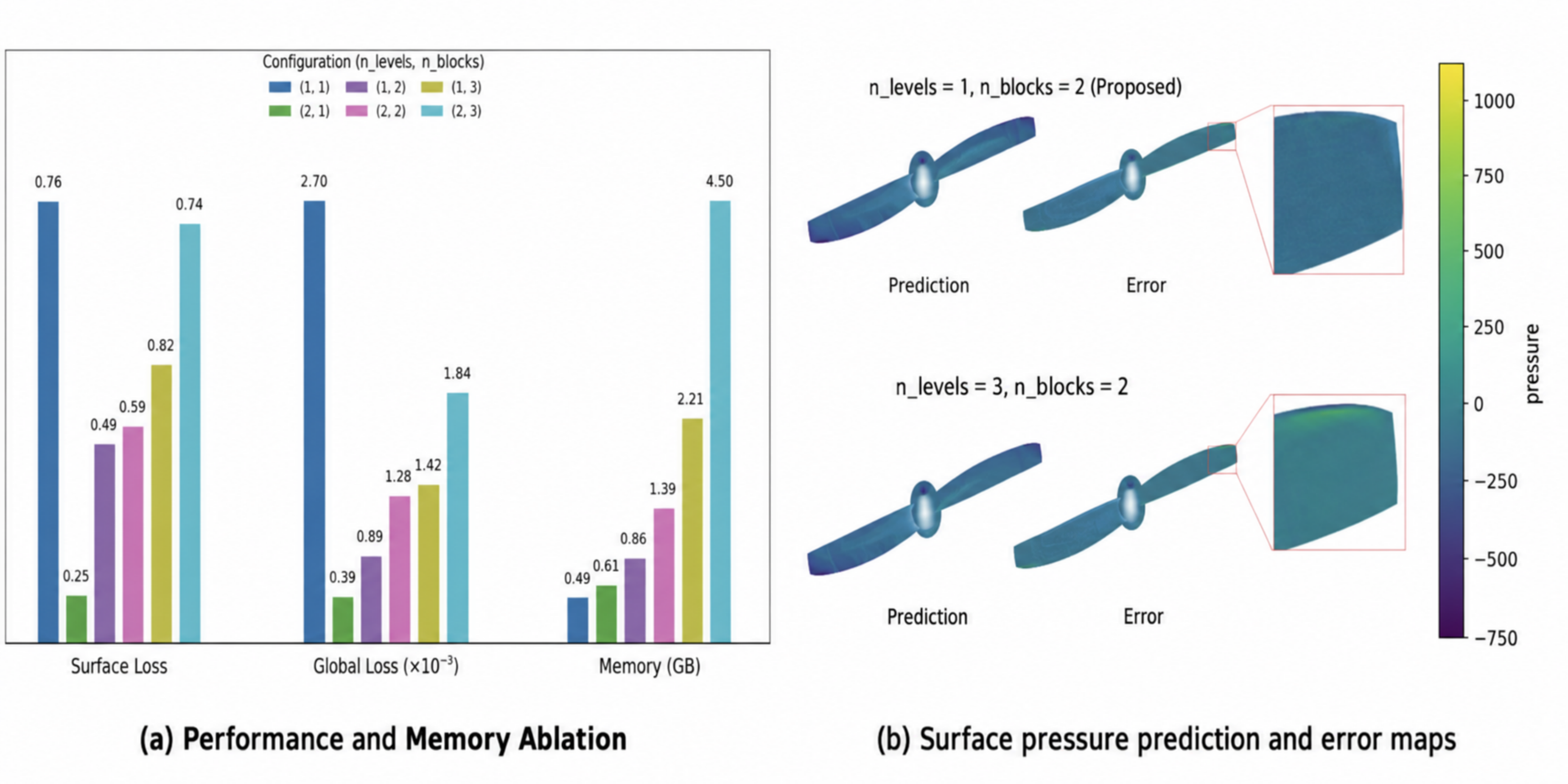}
    \caption{
    \textbf{Quantitative and qualitative ablation of U-Net hierarchy.} 
    (a) Comparison of surface and global losses and GPU memory usage across six configurations of $n_{\text{levels}}$ and block counts ($n_{\text{up}}, n_{\text{down}}$). 
    (b) Example predictions and error maps for representative models $(1,1)$, $(1,2)$, and $(3,2)$.}
    \label{fig:arch_ablation}
\end{figure}

\paragraph{U-Net Depth and Block Count.}
We investigate how hierarchical depth ($n_{\text{levels}}$) and intra-stage convolutional capacity ($n_{\text{up}}, n_{\text{down}}$) affect performance. Figure~\ref{fig:arch_ablation} presents quantitative comparisons of surface loss, global loss, and GPU memory usage across six configurations, as well as qualitative visualizations of surface pressure prediction and error maps. Across all tested configurations, the $(1,2)$ setting offers the best balance between accuracy and efficiency. Shallower models lack contextual coverage, while deeper hierarchies oversmooth local features and incur unnecessary cost. A shallow hierarchy with richer intra-level blocks therefore provides a more effective inductive bias under limited data.

\paragraph{Decoder and Grid SDF Variants.}
We further evaluate the influence of decoder formulation and grid SDF conditioning.
Three configurations are compared: (1) point convolution decoder with the SDF input, (2) interpolation-based decoder without SDFs, and (3) the proposed model combined with interpolation and SDF. All three configurations are trained using both $\mathcal{L}_\text{rel}$ and $\mathcal{L}_\text{global}$.
The results are shown in Table~\ref{tab:decoder_sdf_ablation}.

\begin{table}[H] \vspace{1mm}
\caption{
\textbf{Ablation on decoder formulation and grid SDF conditioning.} 
Interpolation enhances smoothness and stability, while SDF features improve geometric consistency. } \vspace{-1.5mm}
\label{tab:decoder_sdf_ablation}
\centering
\renewcommand{\arraystretch}{1.1}
\begin{tabular}{lcc}
\toprule
\textbf{Configuration} & \textbf{Surface $\downarrow$} & \textbf{Global $\downarrow$} \\
\midrule
PointConv (w/ Grid SDF) & 45.07\% & $7.67\times10^{-3}$ \\
Interp (w/o Grid SDF) & 26.85\% & $9.72\times10^{-4}$ \\
Proposed & \textbf{24.07\%} & $\boldsymbol{3.43\times10^{-4}}$ \\
\bottomrule
\end{tabular}
\renewcommand{\arraystretch}{1.0}
\end{table}

The interpolation-based decoder substantially improves training stability and surface smoothness compared to the original Point Convolution. Point-based decoding is highly sensitive to irregular neighborhood patterns, which vary sharply across propeller blades and often introduce inconsistent or noisy reconstructions. Our decoder, by contrast, provides a stable and geometry-agnostic mapping whose output changes smoothly with vertex position, making it inherently more robust for transferring latent grid features back to complex surfaces.

Meanwhile, incorporating grid SDF features introduces richer geometric context into the latent representation. Unlike positional encodings that only describe relative coordinates, SDF features offer a continuous measure of distance to the surface, helping differentiate near-surface regions from free space and aligning pressure predictions with blade geometry. This lightweight geometric prior improves physical consistency.

Thus, combining interpolation-based decoding with SDF-aware conditioning yields smoother, more coherent predictions that remain geometrically and physically aligned with the underlying flow structure. 

\paragraph{\textbf{Super-Resolution Inference.}}
To evaluate whether FG-ConvNO inherits the resolution-agnostic behavior introduced by DISCO layers, we train the model with the default factorized grid configuration (3×100×200),(150×2×200),(150×100×4), and evaluate it at inference time with a 1.5× finer resolution (applied only to the uncompressed axes).

The resulting surface error differs by only 0.1\% from the native-resolution evaluation, indicating that DISCO blocks maintain stable behavior across grid scales. This resolution robustness is expected to become even more useful for larger datasets or geometries with stronger multiscale features, where higher-resolution latent grids may capture finer flow variations without sacrificing stability.

We further evaluate mesh-level super-resolution by training with only half of the surface vertices while keeping the test resolution unchanged. The test error increases by only 1\%, which is attributable to the reduced discretization accuracy rather than any systematic degradation of the model, indicating that FG-ConvNO remains stable across different mesh densities and can potentially support curriculum strategies that begin with coarse sampling and gradually increase geometric resolution.

\section{Conclusion}

We proposed the Factorized Geometry Convolutional Neural Operator for propeller CFD prediction under limited data. FG-ConvNO simplifies the Convolution hierarchy and the decoding process, incorporates geometry aware features such as SDFs and normals, introduces a global prediction head for thrust and torque, and replaces resolution dependent CNN kernels with discrete continuous convolutions. These components jointly improve stability, discretization consistency and generalization.

Across experiments, FG-ConvNO achieves the best accuracy on both surface pressure and global thrust prediction, while exhibiting stable training behavior. Ablation studies further show that geometric conditioning, interpolation-based decoding, and balanced hierarchical depth collectively contribute to robust learning in complex, high-curvature geometries. Together, these design choices establish a coherent principle: coupling lightweight architectures with geometry-informed representations yields robust and generalizable CFD surrogates, even under limited data availability.

\textbf{Future Directions.}
Currently, the dataset size remains relatively limited, constraining the full exploration of complex geometric–physical correlations. We intend to produce a significantly larger and more diverse dataset of propeller geometries, boundary conditions, and operation regimes, so we can better evaluate model performance and the extrapolation capability at different levels of generality.

Exploring richer geometric encodings, such as curvature-based descriptors or implicit surface representations, could further improve the model’s understanding of complex 3D structures. Additionally, integrating physics-informed regularization or differentiable solver coupling offers a potential path toward physically constrained and transferable neural operators applicable to a broader class of CFD problems.

In this paper, we consider only steady-state propeller flows and an extension of FIGConv to unsteady or turbulent ones is an interesting future direction. Temporal coupling through recurrent or transformer-based operators could enable dynamic flow prediction and wake evolution modeling, bridging the gap between steady and time-resolved behavior.

Finally, while the presented approach shows promising performance for complex propeller geometries in the small-data regime, future work will involve benchmarking with more canonical aerodynamic and hydrodynamic shapes, such as car bodies and airfoils, to further test generalization and robustness across geometric domains.

\newpage
\section*{Acknowledgments}

This work was supported in part by the Bren Endowed Chair at Caltech, 
NVIDIA research support, and computing resources provided by the Caltech 
High-Performance Computing Center. We also thank NVIDIA for additional 
GPU resources used in this work.

\bibliographystyle{plainnat}
\bibliography{aaai2026}

\clearpage
\onecolumn
\appendix

\hfill

\section{Additional Dataset and Simulation Details}

Each propeller case was simulated using the open-source CFD solver OpenFOAM\textregistered\ under steady RANS conditions with the \texttt{k-$\omega$ SST} turbulence model. The inflow velocity ranged from 3–10~m/s, and the angular velocity $\omega$ was fixed to 4,000~rpm. Meshes contained approximately 100k–150k cells, with refined boundary layers near the blade surface to ensure accurate wall-shear stress resolution. The dataset contains 1,000 cases, with 800 used for training and 200 for testing.

For data preprocessing, all surface vertex coordinates were normalized independently along each spatial dimension using the global bounding box of the entire dataset, mapping each axis range to $(0,1)$. This ensures consistent geometric scaling across different propeller configurations and stabilizes the vertex–grid interpolation during training. Surface pressure values were standardized by Gaussian normalization, where both the mean and standard deviation were calculated over the training set. 

During inference, both model outputs and ground-truth pressures were de-normalized before computing evaluation losses to ensure physical comparability. The same normalization and de-normalization procedures were applied to global quantities such as thrust and torque to maintain consistency between local and integral supervision. \\

\vspace{2mm}

\section{Additional Network Architectures Details}
\label{app:architectures}

We summarize the architectural configurations of all compared models, including the proposed FG-ConvNO, implemented uniformly in PyTorch for fair comparison. All models share the same input normalization and conditioning pipeline, with local pressure supervision used throughout and additional global supervision applied only to the specified FG-ConvNO variants. Although the compared methods differ in how spatial dependencies are modeled, the proposed FG-ConvNO introduces a hybrid vertex–grid interaction that bridges continuous and discretized representations.

\vspace{1mm}

\subsection{Common Design Philosophy.}  
All networks take as input the normalized vertex coordinates (and optional condition scalars) and output the predicted surface pressures. This common evaluation setup facilitates comparison across architectures, while differences in feature processing and supervision are described explicitly for each model.

\vspace{2mm}

\subsection{Model Configurations.}  
Below we provide concise configuration-style summaries for the principal models evaluated in this study.  
Each follows a common notation to facilitate direct comparison.

GraphSAGE serves as a lightweight graph-based baseline using mean aggregation.  Its neighborhood sampling of $[20,10,5]$ provides a moderate receptive field suitable for smooth CFD fields.

\begin{verbatim}
class GraphSAGEModelConfig:
    model_arch: str = "graphsage"
    in_channels: int = 3
    out_channels: int = 1
    hidden_channels: List[int] = [128, 128, 64]
    num_layers: int = 3
    aggr_type: str = "mean"
    normalize: bool = True
    dropout: float = 0.1
    activation: str = "relu"
    num_neighbors: List[int] = [20, 10, 5]
    use_self_loops: bool = True
    residual: bool = True
    global_pool: str = "mean"
    global_head_dim: int = 64
\end{verbatim}

GINO combines graph operators with Fourier-based global mixing layers. The low latent resolution ($24^3$) ensures memory efficiency while retaining expressiveness.

\begin{verbatim}
class GINOModelConfig:
    in_channels: int = 3
    out_channels: int = 1
    in_gno_radius: int = 3
    out_gno_radius: int = 3
    fno_hidden_channels: int = 64
    fno_n_layers: int = 4
    fno_lifting_channel_ratio: int = 2
    latent_resolution: (24, 24, 24)
\end{verbatim}

\hfill

\vspace{2mm}

Transolver++ adopts a transformer-style attention mechanism over discretized 3D slices. This architecture is powerful for capturing long-range dependencies.

\begin{verbatim}
class TransolverPlusModelConfig:
    space_dim: int = 3
    n_layers: int = 5
    n_hidden: int = 256
    n_head: int = 8
    mlp_ratio: int = 1
    act: str = "gelu"
    slice_num: int = 64
\end{verbatim}

\hfill

\vspace{2mm}

FG-ConvNO introduces a hybrid feature interaction mechanism between irregular vertex sets and structured latent grids.  
Unlike conventional message-passing GNNs, FIGConv performs bidirectional vertex–grid convolutions, enabling efficient spatial reasoning in both local and global contexts.

\begin{verbatim}
class FIGConvModelConfig:
    out_channels: int = 1
    kernel_size: int = 5
    hidden_channels: [96, 128]
    num_levels: int = 1
    vertex_hidden_dim: int = 96
    feature_dim: int = 3
    feature_hidden_dim: int = 32
    condition_dim: int = 1
    condition_hidden_dim: int = 32
    grid_extra_dim: int = 1
    grid_extra_hidden_dim: int = 64
    pos_encode_dim: int = 32
    num_down_blocks: int = 2
    num_up_blocks: int = 2
    radius_cutoff: float = 0.1
    mlp_channels: [256, 256]
    neighbor_search_radius: float = 1.71
    to_point_sample_method: "interp"
    geometry_fusion_type: "weighted_sum"
\end{verbatim}

\clearpage

\hfill

\subsection{Latent Grid Decomposition.}  
The latent grid configuration for all FIGConv variants followed a factorized anisotropic design to align with the spatial characteristics of propeller flows.  
Specifically, the resolution–format pairs were defined as (3, 100, 200), (150, 2, 200) and (150, 100, 4), representing three orthogonally oriented grids, each emphasizing two high-resolution axes and one compressed axis. This anisotropic decomposition reflects the physical density distribution of the flow, which is denser near the blades and more uniform along the rotation axis, enabling the model to allocate more representational capacity to regions with higher flow complexity. Empirically, this setup achieves an optimal trade-off between spatial expressiveness and computational cost, effectively preserving local vortical structures while maintaining global coherence. \\ 

\subsection{Vertex–Grid Encoder and Neighborhood Search.}  

Feature exchange between the vertex domain and the latent grids is implemented through a continuous kernel-based formulation.  
For each grid center $p_{m,ijk}$ in grid $F_m$, the encoded feature is
\[f_{m,ijk} = 
\sum_{v \in \mathcal{N}(p_{m,ijk})}
\psi_m\big(\phi(v), \phi(p_{m,ijk}), \phi(v - p_{m,ijk})\big)\]
where $\mathcal{N}(p_{m,ijk})=\{v\mid\|v-p_{m,ijk}\|_2 < r_\text{search}\}$ defines a spatial neighborhood of radius $r_\text{search}$.
Setting $r_\text{search}=1.71$ allows each grid cell to access all vertices within the surrounding $2\times2\times2$ voxel cube, which corresponds to roughly eight adjacent cells, achieving smooth yet localized aggregation.
This radius offers an empirically and physically grounded balance between spatial coverage and locality: smaller radii produce fragmented representations, while larger values over-smooth high-frequency flow features. \\

\section{Training and Optimization Settings}
\label{app:training}

All models were trained under identical optimization schedules for consistent comparison.  
The optimizer was Adam with an initial learning rate of $10^{-3}$, decayed by a factor of $0.2$ every $25$ epochs using the \texttt{StepLR} scheduler.  
Weight decay was set to $10^{-5}$.  
Each model trained for $200$~epochs with batch size~$1$ on a single NVIDIA~A100~(32GB) GPU.  
The global supervision weight $\lambda_g$ for integral losses was fixed to~$1.0$.  
Training typically required $10$–$12$~hours per model depending on architecture complexity.

For FG-ConvNO with global supervision, the local and global objectives were optimized jointly:
\[
\mathcal{L} = 
\mathcal{L}_\text{local}
+ \lambda_g\,\mathcal{L}_\text{global},
\]
where $\mathcal{L}_\text{local}$ denotes the relative $L_2$ loss on surface pressure predictions, and $\mathcal{L}_\text{global}$ denotes the mean-squared error on normalized global quantities (e.g., thrust and torque). This hybrid objective stabilizes training and ensures consistency between local flow reconstruction and global aerodynamic performance.

Data augmentation was applied by random vertex subsampling (ratio 0.6–0.8) to improve robustness under sparse or noisy geometric sampling. All experiments were repeated with three random seeds.

\end{document}